# Comparing LSTM and BLSTM Deep Neural Networks for Power Consumption Prediction: Preliminary studies

Davi Guimarães da Silva [a,b,c], Anderson Alvarenga de Moura Meneses [a,c]

[a] Federal University of Western Pará – Graduate Program in Society, Nature and Development

R. Vera Paz, s/n, Salé, CEP 68.035-110, Santarém, PA, Brazil

[b] Federal Institute of Education, Science and Technology of Pará

[c] Federal University of Western Pará, Institute of Geosciences and Engineering, Laboratory of Computational Intelligence

# Comparing Long-Short Term Memory (LSTM) and Bidirectional LSTM Deep Neural Networks for Power Consumption Prediction

**Abstract**

Electric consumption prediction methods are investigated for many reasons such as decision-making related to energy efficiency as well as for anticipating demand in the energy market dynamics. The objective of the present work is the comparison between two Deep Learning models, namely the Long Short-Term Memory (LSTM) and Bi-directional LSTM (BLSTM) for univariate electric consumption Time Series (TS) short-term forecast. The Data Sets (DSs) were selected for their different contexts and scales, aiming the assessment of the models' robustness. Four DSs were used, related to the power consumption of: (a) a household in France; (b) a university building in Santarém, Brazil; (c) the Tétouan city zones, in Morocco; and (c) the Singapore aggregated electric demand. The metrics RMSE, MAE, MAPE and R2 were calculated in a TS cross-validation scheme. The Friedman's test was applied to normalized RMSE (NRMSE) results, showing that BLSTM outperforms LSTM with statistically significant difference ($p = 0.0455$), corroborating the fact that bidirectional weight updating improves significantly the LSTM performance concerning different scales of electric power consumption.

**KEYWORDS:** Electric Consumption Forecast. Deep Learning. Univariate Time Series. Deep Neural Networks. Long-Short Term Memory.

# 1. Introduction

In several operations electric load forecasting has great importance. For example, short-term load forecasting is necessary for the stability of power systems as well as for optimal dispatching (Caro et al., 2020). Also forecast errors imply profit reduction in competitive electricity markets (Bunn, 2000). In the context of energy efficiency (Martinez et al., 2019), electricity consumption forecast is also useful in detecting anomalies for example in end-users behaviors or faulty appliances (Himeur et al., 2021) as well as for avoiding losses and electric energy waste (Cheng et al., 2022). In this sense the application of new technologies represents a great opportunity. For example, in 2022 the Federation of Industries of Santa Catarina State (FIESC), from Brazil, reported that the electric energy waste in Brazil is correspondent to 43 TWh per year (FIESC, 2022), as estimated by the CELESC generation, transmission and distribution company (Centrais Elétricas de Santa Catarina S.A.). Such annual waste is estimated to be that of 20 millions of Brazilian residences.

Therefore, the development of electric power monitoring systems is extremely important for energy efficiency. In order to cope with energy efficiency challenges in planning, distribution and consumption, several Artificial Intelligence (AI) techniques have been investigated (Ahmad et al., 2022), and the application of Deep Neural Networks (DNNs; Chollet, 2018) for consumption prediction is particularly interesting, since those models have advantages such as capacity of generalization for a massive amount of data (big data) and fast response once a model is trained, with possibility of application within Internet of Things (IoT) systems (Serpanos and Wolf, 2018), using cloud or edge platforms. For example, Lee et al. (2019) proposed a system with energy consumption prediction based on the Long-Short Term Memory (LSTM) DNN (Hochreiter e Schmidhuber, 1997) using an IoT system with edge computing for collecting real Time Series (TS) data from an office environment. Also, Da Silva et al. (2022) developed an AI prediction module for an IoT energy consumption monitoring framework and, for different TS data sets, comparing Extreme Gradient Boosting (XGBoost) and Random Forest (RF) algorithms to the LSTM network, which outperformed the previous methods.

The LSTM network is a type of deep Recurrent Neural Network (RNN). RNNs are networks with one or more feedback loops for temporal processing, with two basic uses, related to associative memories and input-output mapping networks (Haykin, 2009), with applications for example in nonlinear prediction and speech processing. Back in the 1990s RNNs were trained for example with Back-Propagation Through Time (BPTT; Rumelhart et al., 1996) or Real Time Recurrent Learning (RTRL; e.g., Robinson and Fallside, 1987; Schmidhuber, 1992), and research on RNNs focused on the relationship between architecture, training and *short-term* memory structures such as tapped-delay line memory and Gamma memory (see Mozer, 1994; Haykin, 2009). Nevertheless, deeper RNNs are not suitable for training with backpropagation due to errors that rapidly decrease (see *Vanishing Gradients in Recurrent Networks*, in Haykin, 2009; p. 795). Vanishing gradients impairs or blocks the network's learning, since small changes due to distant inputs in time may not influence the learning, therefore making long-term dependencies difficult to learn. This is a condition also known as the *Fundamental Deep Learning Problem* (Schmidhuber, 2015).

Thus, the LSTM network was designed to cope with the vanishing gradient problem. The LSTM overcomes this problem due to its structure, which is the same as the traditional RNN, but with memory blocks replacing summation units in the hidden layer and in a broad sense, those blocks are recurrently connected subnets (Graves, 2012). LSTM was able to solve problems in which long term memory is necessary such as context free languages (Gers and Schmidhuber, 2001) and protein secondary structure prediction (Chen and Chaudhari, 2005).

Schuster and Paliwal (1997) explored the possibility of using past and future information in sequence prediction. The basic idea of Bi-directional RNNs (BRNNs) is assessing context in both directions: whereas one RNN processes the sequence data in one direction, another RNN processes the data in the opposite direction. BRNNs were also applied to secondary protein structure prediction (Baldi et al., 1999). The bi-directional scheme applied to the LSTM networks results in the Bi-directional LSTM (BLSTM) networks. Graves and Schmidhuber (2005) applied BLSTM to the framewise phoneme classification task, outperforming other architectures. Graves et al. (2008) applied BLSTM to unconstrained handwritten recognition.

Successful integration of LSTM to IoT systems for consumption prediction are reported in the literature (e.g., Lee et al., 2019; Da Silva et al., 2021) and other investigations show the LSTM ability of outperforming other models in the electric consumption domain (e.g. Schirmer et al., 2019; Da Silva et al., 2022). Besides, the relevance and competitiveness of LSTM and BLSTM models was demonstrated in several energy consumption prediction scenarios (e.g. Fernández-Martínez and Jaramillo-Morán, 2022; Das et al., 2020). Nevertheless, the growing complexity in DNN architectures and the specificities of real-life contexts and problems in power consumption prediction lead to investigations on the robustness of the prediction algorithms on a variety of data sets. In this way, investigation is needed in order to provide statistical analysis regarding a comparison between such models over multiple data sets. Therefore, a research question can be formulated, regarding the performance of such models: Is there statistically significant difference between LSTM and BLSTM over multiple univariate energy consumption data sets?

During the investigation, other factors must also be considered. It is desirable that the time-dependency in the time-series be addressed accordingly. And besides, for a fair comparison due to the nature of AI supervised training, it is necessary, for example, to select a hold-out data subset for performance evaluation (i.e., a data subset which was not used during training).

Therefore, in the present work we compared two DNN architectures for univariate energy consumption TS, namely LSTM network (Hochreiter e Schmidhuber, 1997) and BLSTM (Graves and Schmidhuber, 2005), addressing the aforementioned issues. Thus, firstly four data sets were used, with different characteristics and scales (household, building, city zones, and insular country scales). Secondly the results were evaluated according to a Time Series Cross Validation scheme (TS-CV; Hyndman and Athanasopoulos, 2018), for addressing the time dependency of the predictions. Thirdly, the evaluation of the networks was performed in a hold-out data set consisting of the last month of data for each data set, using the Friedman test (Friedman, 1937) for comparison of models over multiple data sets (Demšar, 2006; García et al., 2010).

Thus, the main contributions of the present paper are: (i) prediction of multiple electric consumption TS data sets, with different scales and characteristics; (ii) a comparison based on a complete methodology for TS prediction regarding TS-CV, hold-out subsets, and the statistical test between LSTM and BLSTM networks' results; (iii) a baseline for future univariate electric consumption TS prediction investigation considering other DNN models.

The remaining of the present article is organized as follows. Section 2 presents the related work. In section 3 the theoretical background is described. The methodology is described in Section 4. The results are presented in Section 5 and discussed in Section 6. Finally, the concluding remarks are presented in Section 7.

2. **Related Work**

Table 1 presents the summary of the main articles related to the present work.

**Table 1.** Related works on DNNs applied to electric consumption prediction.

| Author | Year | Model | Description |
|---|---|---|---|
| Lee et al. | 2019 | LSTM | The authors detected abnormalities and predicted energy consumption in an office through the integration of edge computing and an LSTM network. |
| Kaur et al. | 2019 | LSTM | A unified LSTM scheme was presented for energy management in smart grids, for analyzing and extracting energy patterns aiming demand, forecast and peak reduction. |
| Schirmer et al. | 2019 | LSTM | Regression methods were evaluated for residential energy consumption forecast. LSTM overperforms all the algorithms tested. |
| Mellouli et al. | 2019 | LSTM | Four DNN architectures were evaluated for indoor temperature and energy consumption prediction in a cold room. Stacked LSTM is pointed out as the most efficient DNN among the models tested. |
| Das et al. | 2020 | LSTM, GRU and BLSTM | LSTM, BLSTM and GRU were compared for electric load prediction. BLSTM and GRU performed better in longer prediction horizons. |
| Hou et al. | 2021 | LSTM | LSTM network was applied for residential consumption prediction with load aggregation in selected household. The method proposed reduces the MAPE in the prediction, being useful for microgrid applications. |
| Hadri et al. | 2021 | LSTM, XGBoost and SARIMA | LSTM, XGBoost, and SARIMA algorithms were evaluated for very-short load forecasting in one data set using three strategies (univariate, multivariate, and multistep). None of the algorithms outperforms the others in all cases. However, LSTM reaches the worst results in all cases. |
| Ozer et al. | 2021 | LSTM | LSTM network was used with transfer learning for load forecasting. The authors applied cross-correlation between time series in order to decide which data set should be used for training a model and transfer learning for prediction of a new data set. |
| Da Silva et al. | 2022 | LSTM, RF and XGBoost | LSTM, RF, and XGBoost were evaluated for consumption prediction in three time series with TS-CV and statistical tests. LSTM achieved the best results in both data sets (respectively with $p = 0.0718$ and $p < 0.0001$). A third data set was used as a test of generalization, and was forecasted with models trained with the previous time series. In this case, LSTM outperformed the other algorithms ($p < 0.0001$). |

| Fernández-Martínez and Jaramillo-Morán | 2022 | LSTM and GRU | LSTM and GRU networks, with and without Empirical Mode Decomposition (EMD) and Complete Ensemble EMD (CEEMD) preprocessing, were used for day-ahead power consumption univariate and multivariate forecasting in a hospital. LSTM and GRU models achieved similar performances, but the inclusion of EMD and CEEMD consistently improved the results for the multivariate case. |
|---|---|---|---|
| Shin and Woo | 2022 | LSTM, RF, and XGBoost | LSTM, RF, and XGBoost were evaluated for the energy consumption prediction in Korea. Those algorithms were applied to time series prior and after the COVID-19 pandemic. The best results were achieved by LSTM in the first period and RF in the second period. |
| Shaqour et al. | 2022 | DNN, BI-GRU-FCL, GRU-FCL, BLSTM-FCL, and CNN | Prediction of energy consumption aggregation data was performed by DNN, BI-GRU-FCL, GRU-FCL, BLSTM-FCL, and CNN networks. DNN achieved the higher MAPEs for most aggregation levels and BI-GRU-FCL achieved lower RMSEs. |
| Mubashar et al. | 2022 | LSTM ARIMA | LSTM, ARIMA, and Exponential Smoothing were compared for short-term load forecasting, using real-world data from twelve households. MAE results show that LSTM outperformed the other two methods. |

Lee et al. (2019) proposed a system with energy consumption prediction based on a LSTM network integrated with edge computing. For this purpose, energy consumption data were collected during four months in an office for training and the analysis was conducted in two months data. Two forecast models were implemented (hour and day-ahead). For the hour-ahead forecast, the energy consumption did not present a satisfactory precision. Conversely, for the day-ahead forecast, the training errors were reduced as the number of iterations were increased. Besides a 27% Root Mean Square Error (RMSE), abnormalities could also be detected and the analysis of the electric consumption could be performed.

Kaur et al. (2019) developed a unified scheme based on RNN-LSTM networks for a Smart Grid system, for an integrated approach and data analysis with greater precision. The case study used data from 112 smart homes. The data were pre-processed and decomposed using the High-Order Singular Value Decomposition (HOSVD) for dimensionality reduction. Then, the three network models were applied, as well as the RNN, the RNN-LSTM and Autoregressive Integrated Moving Average (ARIMA). The authors used the metrics RMSE and the Mean Absolute Percentage Error (MAPE). Results indicate that RNN-LSTM obtained the lowest errors (RMSE = 3.35 and MAPE = 5.21%), The RNN (RMSE = 4.613 and MAPE 17.312%) and ARIMA (RMSE = 4.27 and MAPE 29.18%).

Schirmer et al. (2019) evaluated the performance of regression methods for energy consumption prediction, namely: Linear Regression (LR), Decision Trees (DTs), Neural Networks, RNNs, Gated Recurrent Unit (GRU), and LSTM. For that purpose, the authors used the Smart Meters in London data set, which contains data from 5.567 households in London, obtained between November, 2011, and February, 2014. It was demonstrated that LSTM network overperforms all the other algorithms, reducing MAE in up to 26,7% when compared to LR.

Mellouli et al. (2019) presented an approach which uses four DNN architectures for inner temperature and energy consumption in a cold room: LSTM, *Convolutional* LSTM, *Stacked* LSTM, and BLSTM. The data set was divided in five use cases for training and testing. The results pointed out the *Stacked* LSTM as the most efficient for the data used.

Das et al. (2020) described the usage of LSTM network for Miscellaneous Electric Loads (MEL), and compared it to BLSTM and GRU networks. The time series was obtained from an office with capacity for six graduate students in Abu Dhabi, UAE. The data set ranges from April to November, 2017. According to the metrics RMSE and MAE, the three models reached good results depending on the device and the horizon prediction. The authors concluded that the BLSTM network is the most stable model, one day and one week ahead forecast.

Hou et al. (2021) applied the LSTM network to an adaptive load aggregation method for residential consumption. For this purpose, they used a residential load set from the Smart Grid Smart City (SGSC) project to predict the short-term residential load for 50, 100, 150 and 200 randomly selected households, in the period from March 1, 2013 to March 31, 2013. March 2013. The proposed method based on LSTM was compared with the load forecasting method based on SVR and another one based on BPNN. The results showed that when the total load is predicted directly, the MAPE of LSTM equals only 9.1%, which is lower than traditional methods. When the aggregate load is predicted separately, the MAPE of the LSTM is equal to 8.3%, while the MAPE of the SVR-based method is equal to 11.2% and the MAPE of the BPNN-based method is equal to 10.2% . In all cases the proposed method obtains the best MAPE value.

Hadri et al. (2021) investigated the algorithms XGBoost, LSTM, and Seazonal Autoregressive Integrated Moving Average (SARIMA), in order to evaluate empirically the precision of the prediction, the execution time, and the computational complexity. For this purpose, the Dutch Residential Energy Database (DRED) was used, which contains electric energy consumption data in aggregated and disaggregated levels, as well as data related to occupancy, indoor temperature, and weather. The authors conclude that none of the algorithms outperforms the others for the three prediction strategies (univariate, multivariate and multistep). For example, XGBoost outperforms the other algorithms in univariate and multistep cases. ARIMA presented the best performance for the multivariate case. The LSTM model presented the worst results in the three strategies. Thus, XGBoost was chosen for implantation in a IoT platform for real time electricity consumption prediction, because it showed showed the better tradeoff in terms of accuracy and computational cost.

Ozer et al. (2021) used the LSTM network and Cross Correlation (XCORR) in a transfer learning scheme. After data normalization, XCORR is applied between a data base and data from a building , to be predicted, in order to determine which data is more appropriate for training. Then an LSTM model is trained and the transfer learning is performed for another LSTM, for the target building data prediction. RMSE, MAPE, and MAE showed that the LSTM with transfer learning succeeded in comparison to RF, XGB, and Light Gradient Boosting Machine (LGBM) models.

Da Silva et al. (2022) evaluated the algorithms LSTM, XGBoost, and RF, with TS-CV in two data sets. The LSTM network had better performance with a tendency of lower results of RMSE in one data set (UCI-household data set; $p = 0.0718$) and statistically significant difference in the other data set (Santarém, Brazil; LABIC-Building data set; $p < 0.0001$). In a third data set (Singapore data set) used for assessing the capacity of generalization, the forecasting was performed with models trained with the previous data sets. In this case, LSTM also achieved the best results ($p < 0.0001$).

Fernández-Martínez e Jaramillo-Morán (2022) presented a power consumption forecast approach for a medical assistance building using LSTM and GRU networks, testing the *Empirical Mode Decomposition* (EMD) as well as the Complete Ensemble EMD (CEEMD). The data set contains power consumption and meteorological variables collected from September, 2016 to July, 2021. Case studies were conducted for predicting the hourly consumption: a univariate scenario (that takes into account only the whole data set of active energy consumption) and a multivariate scenario (with active power, reactive power, temperature and humidity time series). Results showed that LSTM and GRU networks with EMD and CEEMD preprocessing outperformed the models without preprocessing in all cases. The best results were obtained by LSTM with preprocessing in the multivariate scenario (MAPE = 3.51% and RMSE = 55.06).

Shin and Woo (2022) compared the algorithms XGBoost, LSTM, and RF electric consumption forecast in South Korea dividing the data set in two periods: before and after the COVID-19 pandemics (respectively, period 1 and 2). Data of the period ranging from January 1996 and June 2021 were used for analysis, based on the Total Energy Supply (TES), which also contains highly correlated variables for analysis of the predictions. Results showed that the LSTM model achieved lower RMSE and MAPE in period 1, whereas RF presented lower RMSE e MAPE in period 2. XGB presented the higher errors in the predictions.

Shaqour et al. (2022) used aggregated power consumption and DNN, Bi-directional Gated Recurrent Unit with Fully Connected Layers (BI-GRU-FCL), Gated Recurrent Unit with fully connected layers (GRU-FCL), Long Short-Term Memory with Fully Connected Layers (LSMT-FCL), and Convolutional Neural Network (CNN), for prediction. Results show that DNN reached the higher MAPE for most aggregation levels and Bi-GRU-FCL indicated lower RMSE with a 15% faster training and 40% less parameters regarding DNN.

Mubashar et al. (2022) compared short-term load forecasting performed by LSTM, ARIMA, and Exponential Smoothing using real-world a three months TS from twelve households. MAE results show that LSTM overperformed the other two methods.

As mentioned before, the main contribution of present work is different from previous works since LSTM and BLSTM models were trained, tested and statistically evaluated considering four energy consumption data sets, with different characteristics, magnitudes and locations. The present results may establish a baseline for future tests regarding other networks architectures.

3. **Theoretical Background**

3.1. Long-Short Term Memory (LSTM) Deep Neural Networks

The LSTM DNN architecture was introduced by Hochreiter e Schmidhuber (1997), proposing a solution for the vanishing gradient problem, an usual occurrence in RNNs. In a broad sense, LSTM networks have great capacity for retaining past information (long-term memory), whereas keeping the relevance of recent states (short-term memory).

A common LSTM unit is composed by gates (input, forget, and output). The unit is responsible for the network memory, represented by the activation of the weighted sum, and the gates are means of allowing the flow of information, or not (HOCHREITER and SCHMIDHUBER, 1997; SCHMIDHUBER, 2015).

A LSTM is based on the RNN, however the inner modules have different components. In a LSTM, the cell state is the most important, and aims to transmit information along the network. The information in the cell state is either discarded or modified by the gates. For more details about the LSTM architecture see Graves (2012) and Hasan et al. (2019).

Eqs. 1-6 are used in the gates. (where $W_f$, $W_i$, $W_c$ e $W_o$ are weights, $b_f$, $b_i$, $b_c$ e $b_o$, are bias vectors, $\tilde{C}_t$ is the cell memory, and $\sigma$ is the sigmoid function), according to Hasan et al. (2019).

The forget gate $f_t$ consists in a sigmoid activation function which is applied to the previous hidden state $h_{t-1}$ and the current input $x_t$ for producing a vector (output) where each element is a value between 0 and 1. This layer "decides" which information will be kept or discarded (represented respectively by the values 0 or 1). The output of the forget gate is given by

$$f_t = \sigma(W_f . [h_{t-1}, x_t] + b_f) \tag{1}$$

The input gate $i_t$ is responsible for which information will be kept in the cell state. The input values are the previous hidden state $h_{t-1}$ and the current input $x_t$. The input gate uses a sigmoid and a hyperbolic tangent (*tanh*) function for deciding which information will be added to the cell state. The sigmoid function determines if the current information is important or not, while the *tanh* function returns a value between -1 and +1. Then, both outputs are multiplied by

$$i_t = \sigma(W_i . [h_{t-1}, x_t] + b_i) \text{ and} \tag{2}$$

$$\tilde{C}_t = tanh(W_c . [h_{t-1}, x_t] + b_c). \tag{3}$$

In the second part of this step, the resulting values are combined for updating the cell state. The information of the cell state are updated with the multiplication of the current cell state and the output of the forget gate. If $f_t$ is 0, the result is also 0 and the value becomes insignificant. Otherwise, if $f_t$ is 1, it is kept. Afterwards, the addition updates the cell state according to

$$C_t = f_t \times C_{t-1} + i_t \times \tilde{C}_t. \tag{4}$$

Finally, the output gate $O_t$ determines the final cell state and also the next hidden state $h_t$. In this gate, the previous hidden state $h_{t-1}$ and current input $x_t$ are the inputs for a sigmoid function and the current cell state $C_t$ is passed by a *tanh* function. Then, the sigmoid output and the tanh output are multiplied to determine what information the hidden layer is going to carry, according to

$$O_t = \sigma(W_o \cdot [h_{t-1}, x_t] + b_o) \text{ and} \tag{5}$$

$$h_t = O_t \times tanh(C_t). \tag{6}$$

The structure of gates used by LSTM networks allows the solution of several problems related to sequential models with dependencies, as is the case of electric consumption univariate time series forecast.

3.2. Bidirectional Long-Short Term Memory (BLSTM) Deep Neural Networks

Schuster e Paliwal (1997) presented the Bi-directional RNNs (BRNNs). Their basic idea is that two independent networks can process the input data in opposed sequences, for data whose start and end are known previously, for example for the phoneme boundary estimation problem (Fukada et al., 1999). One sequence is processed in the usual forward direction (forward state), whereas the sequence is also processed from the end to the beginning, as shown in Fig. 1.

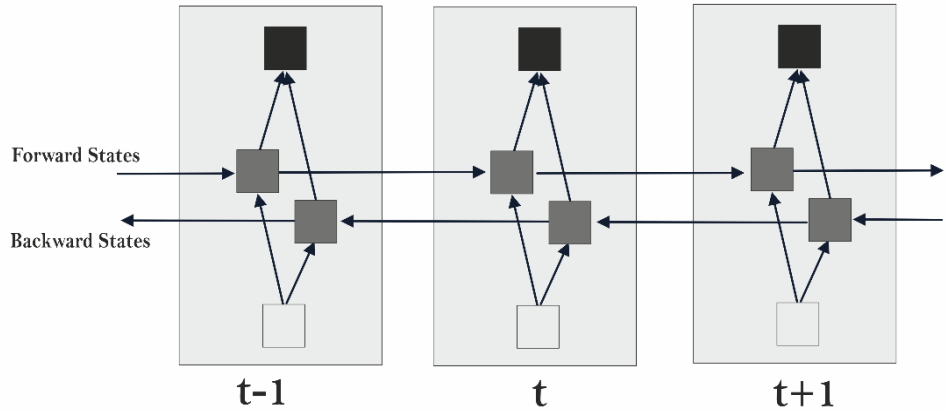

**Fig. 1.** A BRNN cell structure (Schuster and Paliwal, 1997).

Within a BRNN structure the neurons of a regular RNN are divided in a bidirectional form: one for backward states (negative time direction) and other for forward states (positive time direction). The inputs of the inverse direction are not connected to the results of both states. Thus, using two directions of time, input data from the past and from the future can be used. The BRNN concept was combined with the LSTM structure, and successful applications of the BLSTM were developed (e.g., Graves et al., 2008; Graves and Schmidhuber, 2005, 2008).

According to Sharfuddin et al. (2018), in the implementation of a BLSTM, two LSTM layers are used. One of them is responsible for the past states and the other is responsible for the future states, as shown in Fig. 2.

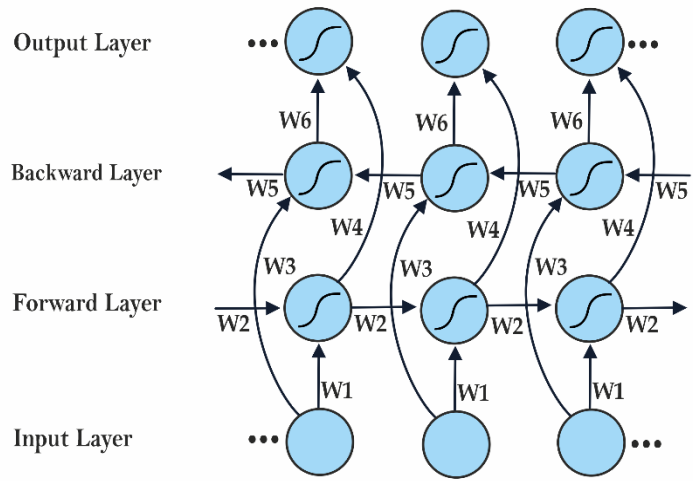

**Fig. 2.** Example of a BLSTM network model (Sharfuddin et al., 2018).

According to Zhao (2018), the structure of BLSTM can be explained as follows: the network has two hidden layers, the horizontal arrows represent the bidirectional flow on the temporal axis; vertical arrows represent unidirectional flow from input layer to hidden layer and from hidden layer to output layer (straight lines). The curved lines are forward and backward unit flows, respectively.

In more detail, Zhao (2018) explains how BLSTM works as: let $S = \{(y^i, x^i)\}_{i=1}^{N}$ represent the set of $N$ samples. For the sample, input $x^i$ has four features: a three-dimensional path and a time clock. The output $y^i$ depends on different tasks. For the hit-miss classification task, $y^i$ has a binary hit-miss value. For the generation task, $y^i$ is the evaluation of the next point $x^{i+1}$. Furthermore, the author points out that a single BLSTM layer can be concatenated with a direct sequence and an inverse sequence, as demonstrated in the following notation:

$$\vec{h}_t = LSTM\,(x_t, \vec{h}_{t-1}) \tag{7}$$

$$\overleftarrow{h}_t = LSTM\,(x_t, \overleftarrow{h}_{t-1}) \tag{8}$$

$$y_t = g\,(W_{\vec{hy}}\,\overleftarrow{h}_t + W_{\overleftarrow{hy}}\,\overleftarrow{h}_t + b_y) \tag{9}$$

where, $LSTM\,(.)$ is used to represent all functions of a standard LSTM, $g\,(.)$ represents the activation function, $W$ represents the weight and $b$ represents the bias of a given layer.

According to Liang et al. (2019), during the bi-directional stage of the network, the output of the forward LSTM-cell sequence is calculated using inputs in the positive direction while the output of the backward LSTM-cell sequence is calculated using the inputs in the reversed direction. the output of the forward LSTM-cell sequence is calculated using inputs in the positive direction while the output of the backward LSTM-cell sequence is calculated using the inputs in the reversed direction. Later the two outputs are then concatenated and placed in a SoftMax function to normalize their values into a probability distribution, which will produce the final output.

In general, regarding the difference between LSTM and BLSTM models, Sharfuddin et al. (2018) state that while LSTM networks allow the inputs in only one direction, BLSTM allows the information flow in both directions, adding a new LSTM layer that inverts the sequence, and the outputs of both layers are combined, for example, with average, sum, multiplication or concatenation. The possibility of two flow directions enables a better learning process.

## 4. Methodology

A conceptual diagram of the methodology is shown in Fig. 3. Each DS was preprocessed for extracting the univariate time series (power and time) and downsample for ten-minutes interval. An Exploratory Data Analysis (EDA) was also performed and the TSs graphs, boxplots, and ACF were obtained (Fig. 4). Then, the data were split for obtaining the TS-CV and the holdout subsets. With the TS-CV data, ten LSTM and ten BLSTM models were generated, which in turn were used for prediction of the last month period in each DS and compared to the holdout data. The metrics RMSE, NRMSE, MAE, MAPE, and R2 were then obtained for tables and boxplots. The metric NRMSE was used for comparison among all DSs.

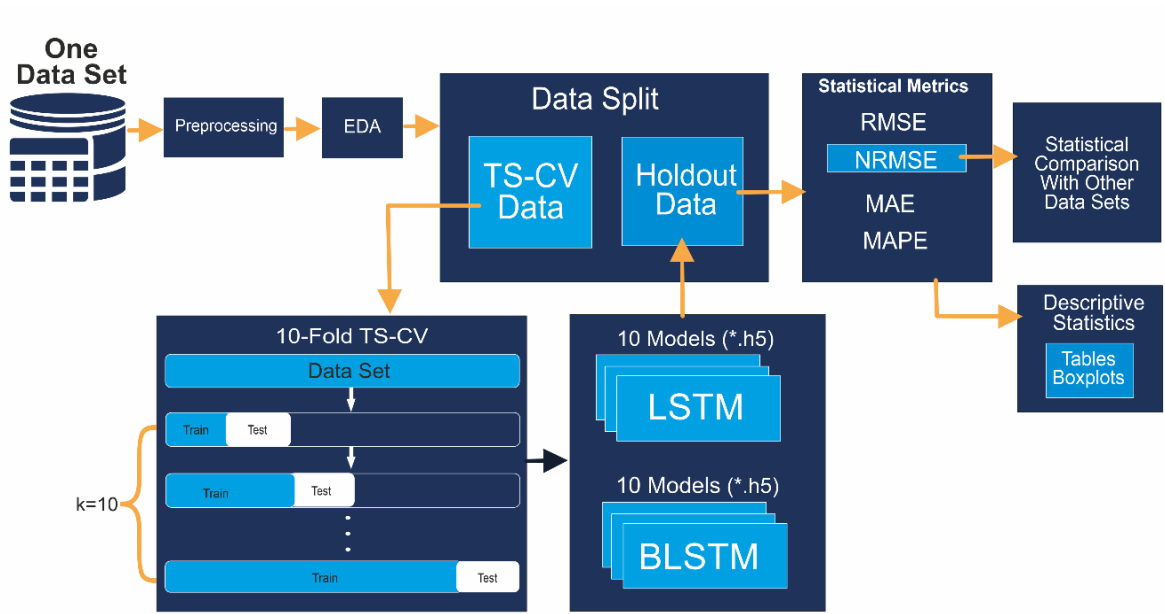

**Fig. 3.** Conceptual diagram of the methodology.

### 4.1 Case study and Data Sets

In the present work, the experiments were conducted for statistical comparison between LSTM and BLSTM networks with TS-CV. The results were evaluated in four data sets: a) the power consumption of an individual household; b) the power consumption of a universitary building with classrooms, bathrooms, a library, an auditorium, cafeteria and administrative departments; c) power consumption in three zones of a city; and d) the power consumption of an insular country, which are described in the next subsections. Table 2 shows a summary of the values for TS-CV and holdout of the data sets.

**Table 2.** Summary of the TS-CV and holdout subsets for each data set.

| Data Sets | Set Train-Test | Set Validation |
|---|---|---|
| UCI-Household | 203801 (98.21%) | 3725 (1.79%) |
| LABIC-Building | 29492 (87.18%) | 4338 (12.82%) |
| Tetouan-Zones | 48096 (91.76%) | 4320 (8.24%) |
| Singapore | 20352 (93.19%) | 1488 (6.81%) |

*4.1.1   UCI-Household Data Set*

The UCI-Household data set is widely used for electric consumption prediction (e.g. Kim and Cho, 2018; Le et al., 2019; Khan et al., 2021). This data set is provided by the *UCI Machine Learning Repository* (Individual household electric power consumption Data Set) and contains data obtained from a household in Sceaux, França. The data were collected from December 2006 to November 2010 (47 months), minute by minute, with a total of 2.075.259 points in kilowatts (kW).

In our investigation the univariate time series of global active power was used, downsampled to 10 minutes intervals (207.526 points). Data from December, 2006 to October, 2010 (203.801 points) were used for TS-CV and data from November, 2010 (3.725 points, 1.79% of the data set) were used as holdout set.

*4.1.2   LABIC-Building Data Set*

The LABIC-Building data set is related to a building of the Federal University of Western Pará (Universidade Federal do Oeste do Pará, UFOPA) with high power demand of AC systems, which is a characteristic of the Amazon region. Data were obtained is Santarém city, Pará State, Brazil (Da Silva et al., 2021, 2022).

The time series contains 256.092 points of aggregated active power in Watts (W). After downsampling the data in 10 minutes intervals, the data set remained with 33.830 points. For the TS-CV, 29.492 points, ranging from January to July 2019, which corresponds to approximately 87% of the data set. The holdout set was composed by 4338 points from August, 2019 (12.82% of the data set).

*4.1.3   Tetouan-Zones Data Set*

The Tetouan-Zones data set is related to the power distribution of three zones in Tetouan city, Morocco and was used in several works for power consumption forecast (e.g. Salam and El Hibaoui, 2018, 2021; Singh et al., 2018). The period corresponds to January to December 2017 (12 months), presented in 10 minutes intervals with a total of 52.416 points in kilowatts (kW).

In our research the period from January to November 2017 (48.096 points) was used for TS-CV, which corresponds to approximately 92% of the data set, and the data points of December 2017 was used as a holdout set (4.320 points, 8.24% of the data set).

*4.1.4   Singapore Data Set*

The Singapore *data set* is a time series of electric power demand in large scale collected in the whole country provided by the website *Energy Market Company Pte Ltd* (EMC; https://www.emcsg.com/marketdata/priceinformation). Its attributes are price type, day, period, price (in $/MW) and demand (MW).

The univariate time series of demand with 30 minutes intervals was used. The time period is from January, 2010, to March, 2011, with 21.840 points. For the TS-CV, 20352 points, ranging from January, 2010, to February 2011, which corresponds to approximately 93% of the data set. The holdout set was composed by 1488 points from March, 2011 (6.81% of the data set).

**4.2  Training and Testing**

The computational resources were a desktop computer with an Intel® Core™ i5 - 7400 processor with a 16GB RAM, as well as an 11GB VRAM Nvidia RTX 2080Ti GPU.

LSTM and BLSTM networks were implemented in Python using Keras 2.6.0, and Tensorflow 2.6.0 as backend.

Due to the time dependency of the information, the training data set must be presented in sequence, containing observations in a time period prior to the test set. Thus, a TS-CV scheme (Hyndman and Athanasopoulos, 2018; Hewamalage et al., 2023) meets that requirement, also providing information on the robustness of the models regarding the length of the training set. The *TimeSeriesSplit* function of the scikit-learn was used (see also Da Silva et al., 2022). In the present work, 10 partitions (folds) for all training sets were used ($k = 10$).

The main parameters of LSTM and BLSTM are described in Table 3.

**Table 3.** LSTM and BLSTM networks' parameters.

| Parameter | Value |
| --- | --- |
| MinMaxScaler | feature_range=(0,1) |
| units | 100 |
| epochs | 100 |
| batch | 32 |
| dropout | 0.3 |
| optimizer | rmsprop |
| Sliding window | 90 |
| activation | linear |
| loss | mse |

For each data set in the TS-CV stage, 10 models were obtained for each architecture (LSTM and BLSTM). The 20 models obtained for each data set were saved in the format ".h5" for prediction of the hold out subset (1 month data), then generating performance metrics for comparison. The average Normalized RMSEs (NRMSE) were used for the Friedman test.

## 4.3 Performance Metrics

Five metrics were used for evaluating the models' performance, namely, RMSE, NRMSE, MAE, MAPE, and the determination coefficient $R^2$. The NRMSE is the metric used for comparison of data sets with different scales with the Friedman's test, since it uses normalization. Such metrics are described by

$$RMSE = \sqrt{\left[\frac{1}{n}\sum_{i=1}^{n}(|y_i - \hat{y}_i|)^2\right]} \quad (10)$$

$$NRMSE = \frac{RMSE}{\bar{y}} \quad (11)$$

$$MAE = \frac{1}{n}\sum_{i=1}^{n}|y_i - \hat{y}_i| \quad (12)$$

$$MAPE = \frac{1}{n}\sum_{i=1}^{n}\left|\frac{y_i - \hat{y}_i}{y_i}\right| * 100 \quad (13)$$

$$R^2 = 1 - \frac{\sum_{i=1}^{n}(y_i - \hat{y}_i)^2}{\sum_{i=1}^{n}(y_i - \bar{y}_i)^2} \quad (14)$$

where $y_i$, $\hat{y}_i$, and $n$ are respectively the measured value, the value predicted and the total number of points; $\bar{y}$ is the mean value of the models' outputs. Hewamalage et al. (2023) provide more details on performance metrics and other topics in forecast evaluation.

### 4.4 Statistical Analysis

The Friedman test was applied (Friedman, 1937; García et al., 2010) for comparing models obtained in the TS-CV scheme. With the Friedman test, it is possible to determine if there is statistical significance difference considering multiple data sets. The number of 10 NRMSE values for each model, in each data set, as suggested by Witten and Frank (2005), makes appropriate a non-parametric test such as Friedman's, which is based on the ranks of the NRMSE results.

The first step of the Friedman test is converting the original results in ranks. Such ranks vary from 1 to *k*, where *k* is the number of models tested. Thus, the *k* models are classified according to each data set *N* separately. For tied scores, an average rank is attributed (see also Demšar, 2006).

The Friedman statistics $\chi_F^2$ is then calculated according to

$$\chi_F^2 = \frac{12N}{K(K+1)}\left[\sum_j R_j^2 - \frac{k(k+1)^2}{4}\right] \quad (15)$$

with $k$-1 degrees of freedom. $R_j$ is the average rank of the models.

The null hypothesis of the test is that there is no statistically significant difference between models. A p-value $p$ is calculated and the threshold 0.05 was adopted. If $p > 0.05$. the null hypothesis is rejected and at least one of them differs from the others. In our case the Nemenyi test was also applied (Nemenyi; 1963; Barrow et al., 2013). The statistical tests were performed in R (https://www.r-project.org/), with the functions "friedman.test" and frdAllPairsNemenyiTest.

## 5. Results

### 5.1 Exploratory Data Analysis

The time series, quarterly boxplots, and ACFs are shown in Fig. 4. The UCI-Household, LABIC-Building, and Singapore data sets were analyzed previously by Da Silva et al. (2022). The Tetouan-Zones data set was included in the present work. The data sets represent different patterns of consumption, contexts, lengths and scales, chosen in order to test the robustness of the algorithms.

As examples of the data sets' characteristics, in UCI-Household data set, in the second and third quarters the power consumption is lower than the first and fourth quarters, due to the electric load related to the winter. In the Labic-Building, the second quarter represents the months of March to May a period with higher consumption of Air Conditioning Systems.

Another noticeable characteristic of the data sets, in the ACF plotted, a small blue region close to the horizontal axis can be noticed. It represents the values with no statistical significance for ACF. In other words, until the lag 100, almost all lags show statistically significant auto-correlations.

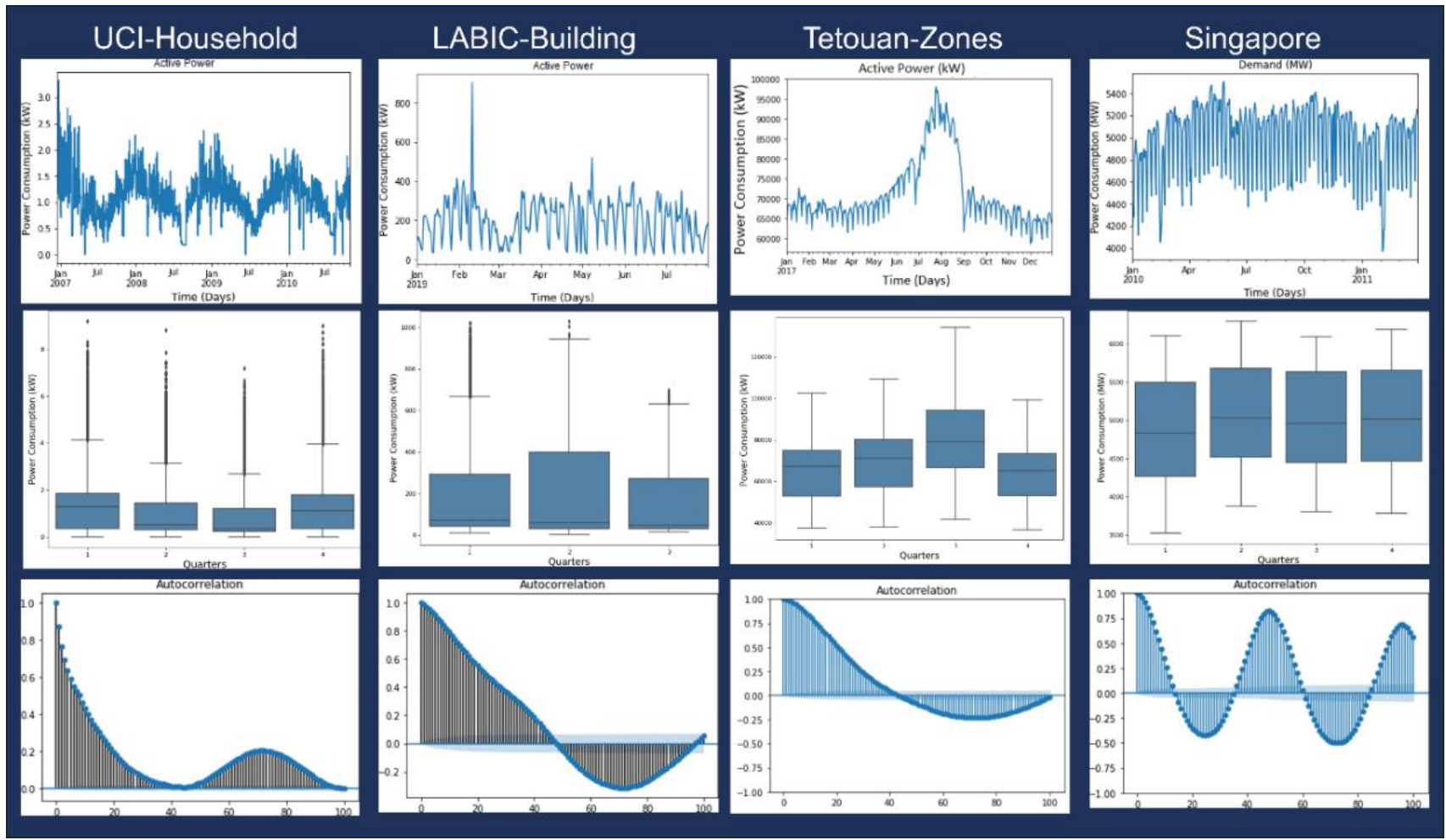

**Fig. 4.** (a) Power consumption time series; (b) Quarterly boxplots; and (c) Auto-Correlation Functions (ACFs) respectively for the four data sets used in the present work.

## 5.2 Computational Results

In the next subsections, the results for the LSTM and BLSTM networks for the UCI-Household, LABIC-Building, Tetouan-Zones, and Singapore data sets, with the holdout data sets. The results of the statistical tests also will be described, as well as the TS-CV execution times.

### 5.2.1 LSTM and BLSTM results for holdout sets

(A) *UCI-Household Data Set*

Tables 4 and 5 show the metrics RMSE, NRMSE, MAE, MAPE, and $R^2$ for 10 models each (obtained by TS-CV), respectively for LSTM and BLSTM models applied to the UCI-Household holdout subset. Fig. 5 shows the NRMSE boxplots for both models. Fig. 6 shows a comparison between typical predictions and the real values of the holdout data set.

**Table 4.** LSTM models' results for the UCI-Household holdout subset.

| LSTM Model | Metrics - LSTM models - UCI-Household holdout | | | | |
|---|---|---|---|---|---|
| | RMSE | NRMSE | MAE | MAPE (%) | R² |
| 1 | 0,484 | 0.065 | 0.272 | 27.3 | 74.4 |
| 2 | 0.481 | 0.064 | 0.266 | 25.6 | 74.7 |
| 3 | 0.477 | 0.064 | 0.275 | 28.9 | 75.1 |
| 4 | 0.594 | 0.079 | 0.394 | 51.2 | 61.4 |
| 5 | 0.489 | 0.065 | 0.270 | 25.4 | 73.9 |
| 6 | 0.620 | 0.083 | 0.414 | 50.9 | 58.0 |
| 7 | 0.548 | 0.073 | 0.349 | 41.9 | 67.2 |
| 8 | 0.476 | 0.064 | 0.270 | 27.1 | 75.2 |
| 9 | 0.480 | 0.064 | 0.268 | 27.4 | 74.8 |
| 10 | 0.605 | 0.081 | 0.407 | 50.6 | 59.9 |
| Average | 0.525 | 0.070 | 0.318 | 35.6 | 69.5 |
| Median | 0.486 | 0.065 | 0.273 | 28.1 | 74.1 |
| St-Dev | 0.060 | 0.008 | 0.065 | 11.6 | 7.1 |
| Min | 0.476 | 0.064 | 0.266 | 25.4 | 58.0 |
| Max | 0.620 | 0.083 | 0.414 | 51.2 | 75.2 |

**Table 5.** BLSTM models' results for the UCI-Household holdout subset.

| BLSTM Model | Metrics - BLSTM models - UCI-Household holdout ||||| 
| --- | --- | --- | --- | --- | --- |
| | RMSE | NRMSE | MAE | MAPE (%) | R² |
| 1 | 0.497 | 0.066 | 0.291 | 30.5 | 72.9 |
| 2 | 0.510 | 0.068 | 0.286 | 27.3 | 71.5 |
| 3 | 0.506 | 0.068 | 0.290 | 27.6 | 72.0 |
| 4 | 0.506 | 0.068 | 0.290 | 27.6 | 72.0 |
| 5 | 0.499 | 0.067 | 0.307 | 36.2 | 72.7 |
| 6 | 0.488 | 0.065 | 0.286 | 31.2 | 74.0 |
| 7 | 0.491 | 0.066 | 0.266 | 24.0 | 73.6 |
| 8 | 0.499 | 0.067 | 0.287 | 24.1 | 72.8 |
| 9 | 0.485 | 0.065 | 0.265 | 23.6 | 74.3 |
| 10 | 0.522 | 0.070 | 0.391 | 59.7 | 70.2 |
| Average | 0.500 | 0.067 | 0.296 | 31.2 | 72.6 |
| Median | 0.499 | 0.067 | 0.288 | 27.6 | 72.8 |
| St-Dev | 0.011 | 0.001 | 0.036 | 10.7 | 1.2 |
| Min | 0.485 | 0.065 | 0.265 | 23.6 | 70.2 |
| Max | 0.522 | 0.070 | 0.391 | 59.7 | 74.3 |

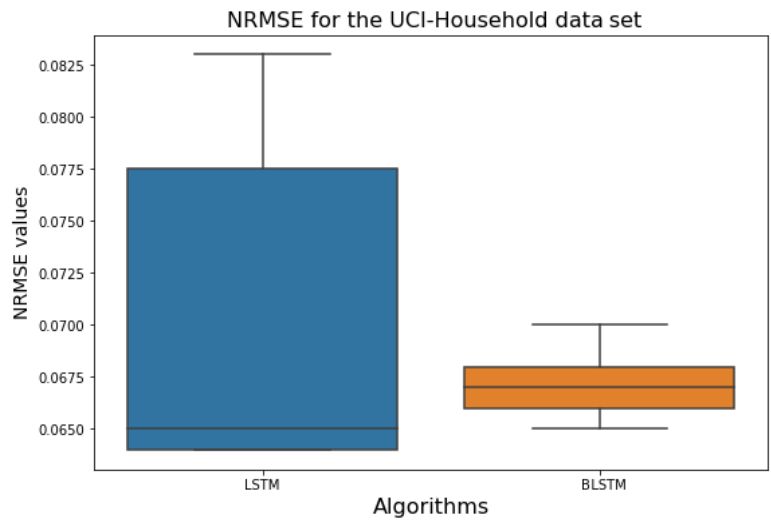

**Fig. 5.** NRMSE boxplots for LSTM and BLSTM models applied to the UCI-Household holdout subset.

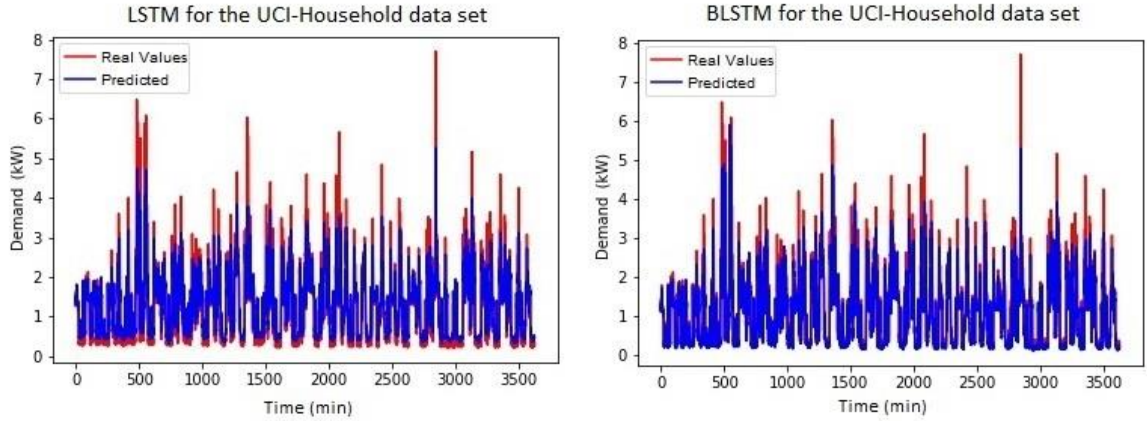

**Fig. 6.** Typical LSTM and BLSTM predictions for the holdout set (UCI-Household data set).

(B) *LABIC-Building Data Set*

Tables 6 and 7 show the metrics RMSE, NRMSE, MAE, MAPE, and $R^2$ for 10 models each (obtained by TS-CV), respectively for LSTM and BLSTM models applied to the LABIC-Building holdout subset. Fig. 7 shows the NRMSE boxplots for both models. Fig. 8 shows a comparison between typical predictions and the real values of the holdout data set.

**Table 6.** LSTM models' results for the LABIC-Building holdout subset.

| *LSTM Model* | **Metrics - LSTM models -** LABIC-Building holdout | | | | |
| --- | --- | --- | --- | --- | --- |
| | RMSE | NRMSE | MAE | MAPE (%) | R² |
| 1 | 35.883 | 0.032 | 18.079 | 12.8 | 97.8 |
| 2 | 36.193 | 0.032 | 22.095 | 24.4 | 97.8 |
| 3 | 37.972 | 0.033 | 18.648 | 11.3 | 97.6 |
| 4 | 35.352 | 0.031 | 17.933 | 14.3 | 97.9 |
| 5 | 35.065 | 0.031 | 17.912 | 13.1 | 97.9 |
| 6 | 37.042 | 0.033 | 22.176 | 22.1 | 97.7 |
| 7 | 35.660 | 0.031 | 18.113 | 13.6 | 97.9 |
| 8 | 36.060 | 0.032 | 19.389 | 15.9 | 97.8 |
| 9 | 36.158 | 0.032 | 20.287 | 17.3 | 97.8 |
| 10 | 34.809 | 0.031 | 17.981 | 13.5 | 98.0 |
| Average | 36.020 | 0.032 | 19.261 | 15.8 | 97.8 |
| Median | 35.971 | 0.032 | 18.380 | 13.9 | 97.8 |
| St-Dev | 0.936 | 0.001 | 1.697 | 4.3 | 0.1 |
| Min | 34.809 | 0.031 | 17.912 | 11.3 | 97.6 |
| Max | 37.972 | 0.033 | 22.176 | 24.4 | 98.0 |

**Table 7.** BLSTM models' results for the LABIC-Building holdout subset.

| BLSTM Model | Metrics - BLSTM models - LABIC-Building holdout | | | | |
|---|---|---|---|---|---|
| | RMSE | NRMSE | MAE | MAPE (%) | R² |
| 1 | 36.587 | 0.032 | 22.003 | 22.6 | 97.8 |
| 2 | 39.469 | 0.035 | 27.645 | 36.5 | 97.4 |
| 3 | 36.895 | 0.032 | 18.588 | 12.4 | 97.7 |
| 4 | 37.574 | 0.033 | 25.741 | 32.1 | 97.6 |
| 5 | 34.093 | 0.030 | 17.225 | 11.9 | 98.0 |
| 6 | 34.116 | 0.030 | 19.217 | 18.5 | 98.0 |
| 7 | 33.521 | 0.030 | 16.427 | 11.0 | 98.1 |
| 8 | 34.342 | 0.030 | 17.134 | 11.2 | 98.0 |
| 9 | 34.413 | 0.030 | 20.092 | 21.1 | 98.0 |
| 10 | 35.989 | 0.032 | 21.705 | 22.0 | 97.8 |
| Average | 35.700 | 0.031 | 20.578 | 19.9 | 97.9 |
| Median | 35.201 | 0.031 | 19.654 | 19.8 | 97.9 |
| St-Dev | 1.925 | 0.002 | 3.746 | 8.9 | 0.2 |
| Min | 33.521 | 0.030 | 16.427 | 11.0 | 97.4 |
| Max | 39.469 | 0.035 | 27.645 | 36.5 | 98.1 |

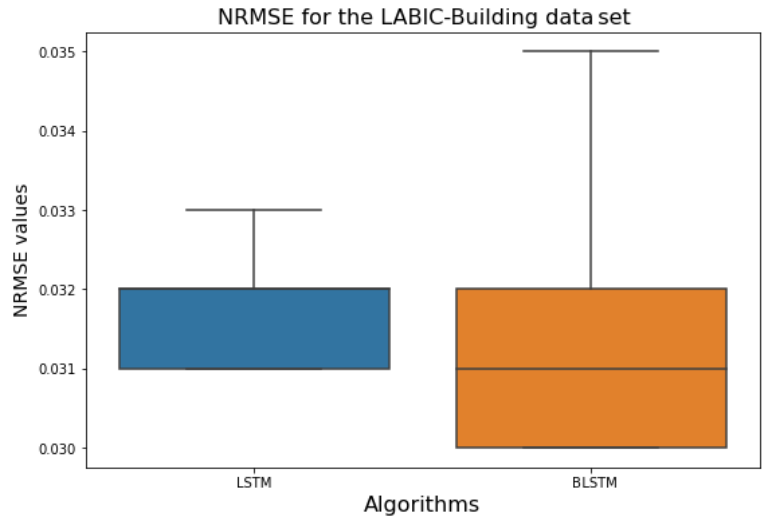

**Fig. 7.** NRMSE boxplots for LSTM and BLSTM models applied to the LABIC-Building holdout subset.

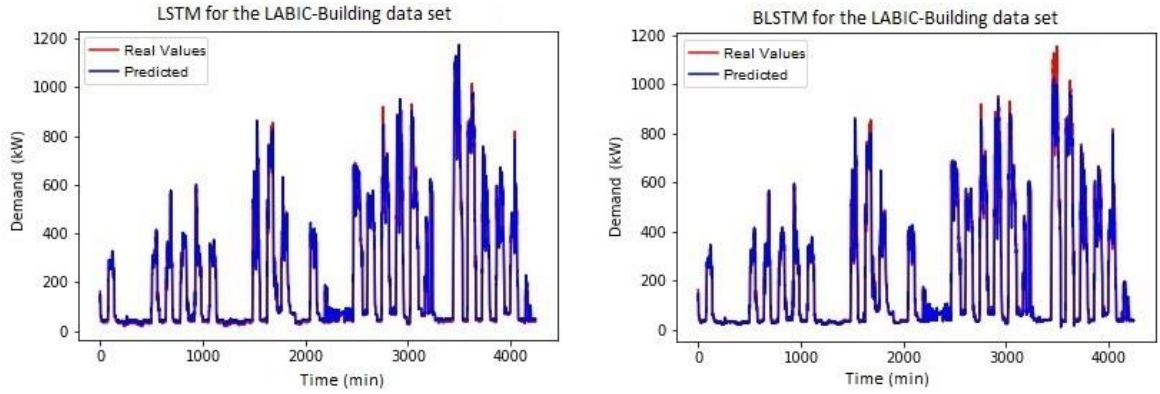

**Fig. 8.** Typical LSTM and BLSTM predictions for the holdout set (LABIC-Building data set).

(C) *Tetouan-Zones Data Set*

Tables 8 and 9 show the metrics RMSE, NRMSE, MAE, MAPE, and $R^2$ for 10 models each (obtained by TS-CV), respectively for LSTM and BLSTM models applied to the Tetouan-Zones holdout subset. Fig. 9 shows the NRMSE boxplots for both models. Fig. 10 shows a comparison between typical predictions and the real values of the holdout data set.

**Table 8.** LSTM models' results for the Tetouan-Zones holdout subset.

| LSTM Model | Metrics - LSTM models - Tetouan-Zones holdout | | | | |
|---|---|---|---|---|---|
| | RMSE | NRMSE | MAE | MAPE (%) | R² |
| 1 | 880.601 | 0.016 | 707.742 | 1.2 | 99.6 |
| 2 | 806.638 | 0.015 | 591.798 | 0.9 | 99.7 |
| 3 | 747.032 | 0.014 | 569.353 | 1.0 | 99.7 |
| 4 | 1264.597 | 0.023 | 1013.582 | 1.5 | 99.2 |
| 5 | 832.836 | 0.015 | 623.614 | 0.9 | 99.7 |
| 6 | 785.838 | 0.014 | 612.084 | 1.0 | 99.7 |
| 7 | 952.206 | 0.017 | 807.024 | 1.4 | 99.6 |
| 8 | 833.170 | 0.015 | 648.251 | 1.0 | 99.7 |
| 9 | 810.916 | 0.015 | 637.257 | 1.1 | 99.7 |
| 10 | 654.841 | 0.012 | 472.449 | 0.8 | 99.8 |
| Average | 856.867 | 0.016 | 668.315 | 1.1 | 99.6 |
| Median | 821.876 | 0.015 | 630.436 | 1.0 | 99.7 |
| St-Dev | 163.161 | 0.003 | 149.273 | 0.2 | 0.2 |
| Min | 654.841 | 0.012 | 472.449 | 0.8 | 99.2 |
| Max | 1264.597 | 0.023 | 1013.582 | 1.5 | 99.8 |

**Table 9.** BLSTM models' results for the Tetouan-Zones holdout subset.

| BLSTM Model | Metrics - BLSTM models - Tetouan-Zones holdout | | | | |
|---|---|---|---|---|---|
| | RMSE | NRMSE | MAE | MAPE (%) | R² |
| 1 | 618.613 | 0.011 | 457.710 | 0.8 | 99.8 |
| 2 | 576.300 | 0.010 | 429.827 | 0.7 | 99.8 |
| 3 | 834.071 | 0.015 | 641.204 | 1.2 | 99.7 |
| 4 | 850.158 | 0.015 | 653.708 | 1.0 | 99.6 |
| 5 | 730.202 | 0.013 | 579.820 | 1.0 | 99.7 |
| 6 | 648.534 | 0.012 | 514.702 | 0.8 | 99.8 |
| 7 | 573.357 | 0.010 | 407.655 | 0.6 | 99.8 |
| 8 | 523.353 | 0.009 | 386.385 | 0.6 | 99.9 |
| 9 | 581.298 | 0.011 | 432.836 | 0.7 | 99.8 |
| 10 | 526.451 | 0.010 | 380.853 | 0.6 | 99.9 |
| Average | 646.234 | 0.012 | 488.470 | 0.8 | 99.8 |
| Median | 599.956 | 0.011 | 445.273 | 0.7 | 99.8 |
| St-Dev | 119.514 | 0.002 | 103.106 | 0.2 | 0.1 |
| Min | 523.353 | 0.009 | 380.853 | 0.6 | 99.6 |
| Max | 850.158 | 0.015 | 653.708 | 1.2 | 99.9 |

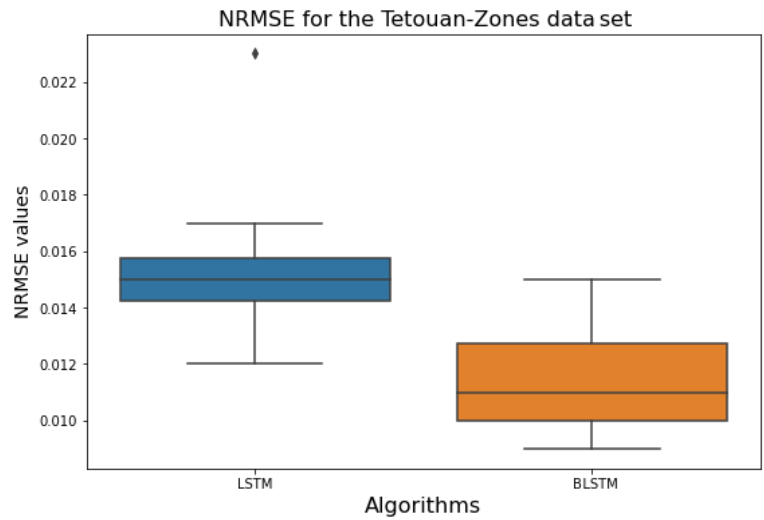

**Fig. 9.** NRMSE boxplots for LSTM and BLSTM models applied to the Tetouan-Zones holdout subset.

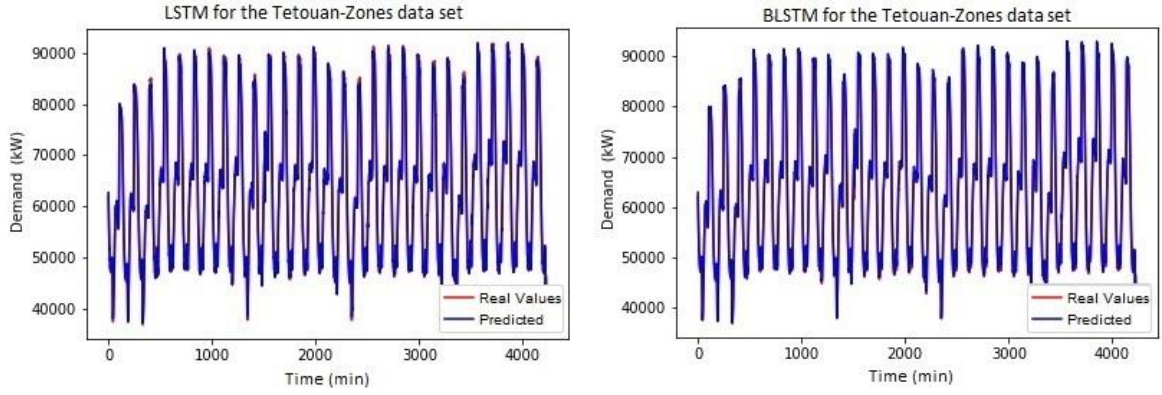

**Fig. 10.** Typical LSTM and BLSTM predictions for the holdout set (Tetouan-Zones data set).

(D) *Singapore Data Set*

Tables 10 and 11 show the metrics RMSE, NRMSE, MAE, MAPE, and $R^2$ for 10 models each (obtained by TS-CV), respectively for LSTM and BLSTM models applied to the Singapore holdout subset. Fig. 11 shows the NRMSE boxplots for both models. Fig. 12 shows a comparison between typical predictions and the real values of the holdout data set.

**Table 10.** LSTM models' results for the Singapore holdout subset.

| *LSTM Model* | **Metrics - LSTM models -** Singapore holdout ||||| 
|---|---|---|---|---|---|
| | RMSE | NRMSE | MAE | MAPE (%) | R² |
| 1 | 35.096 | 0.016 | 28.374 | 0.6 | 99.7 |
| 2 | 35.797 | 0.017 | 28.957 | 0.6 | 99.7 |
| 3 | 30.836 | 0.014 | 24.218 | 0.5 | 99.8 |
| 4 | 42.846 | 0.020 | 35.388 | 0.7 | 99.6 |
| 5 | 32.755 | 0.015 | 26.533 | 0.5 | 99.7 |
| 6 | 48.184 | 0.022 | 39.409 | 0.8 | 99.5 |
| 7 | 31.026 | 0.014 | 24.818 | 0.5 | 99.8 |
| 8 | 46.894 | 0.022 | 40.107 | 0.8 | 99.5 |
| 9 | 36.283 | 0.017 | 28.940 | 0.6 | 99.7 |
| 10 | 29.291 | 0.014 | 23.562 | 0.5 | 99.8 |
| Average | 36.901 | 0.017 | 30.031 | 0.6 | 99.7 |
| Median | 35.446 | 0.017 | 28.657 | 0.6 | 99.7 |
| St-Dev | 6.781 | 0.003 | 6.131 | 0.1 | 0.1 |
| Min | 29.291 | 0.014 | 23.562 | 0.5 | 99.5 |
| Max | 48.184 | 0.022 | 40.107 | 0.8 | 99.8 |

**Table 11.** BLSTM models' results for the Singapore holdout subset.

| BLSTM Model | Metrics - BLSTM models - Singapore holdout | | | | |
|---|---|---|---|---|---|
| | RMSE | NRMSE | MAE | MAPE (%) | $R^2$ |
| 1 | 39.755 | 0.019 | 30.713 | 0.6 | 99.6 |
| 2 | 74.322 | 0.035 | 63.770 | 1.2 | 98.7 |
| 3 | 27.914 | 0.013 | 22.190 | 0.5 | 99.8 |
| 4 | 29.148 | 0.014 | 22.167 | 0.4 | 99.8 |
| 5 | 33.638 | 0.016 | 26.725 | 0.5 | 99.7 |
| 6 | 24.266 | 0.011 | 18.874 | 0.4 | 99.9 |
| 7 | 29.963 | 0.014 | 25.081 | 0.5 | 99.8 |
| 8 | 22.716 | 0.011 | 17.727 | 0.4 | 99.9 |
| 9 | 24.229 | 0.011 | 18.644 | 0.4 | 99.9 |
| 10 | 29.269 | 0.014 | 23.618 | 0.5 | 99.8 |
| Average | 33.522 | 0.016 | 26.951 | 0.5 | 99.7 |
| Median | 29.209 | 0.014 | 22.904 | 0.5 | 99.8 |
| St-Dev | 15.179 | 0.007 | 13.537 | 0.3 | 0.4 |
| Min | 22.716 | 0.011 | 17.727 | 0.4 | 98.7 |
| Max | 74.322 | 0.035 | 63.770 | 1.2 | 99.9 |

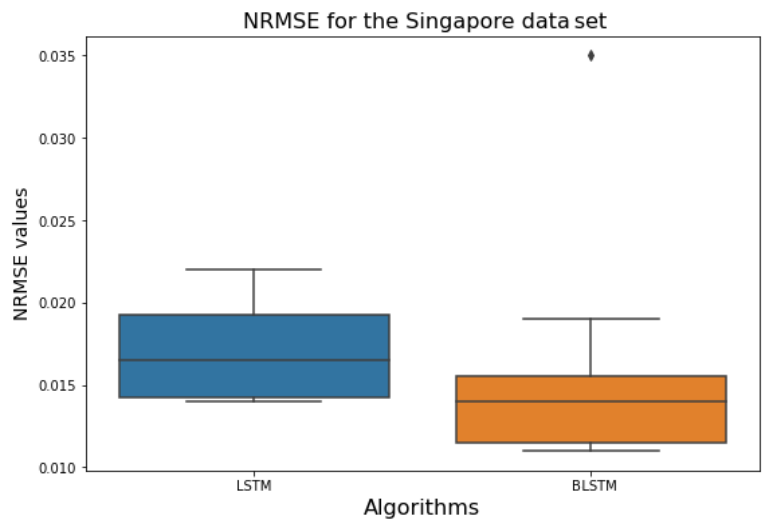

**Fig. 11.** NRMSE boxplots for LSTM and BLSTM models applied to the Singapore holdout subset.

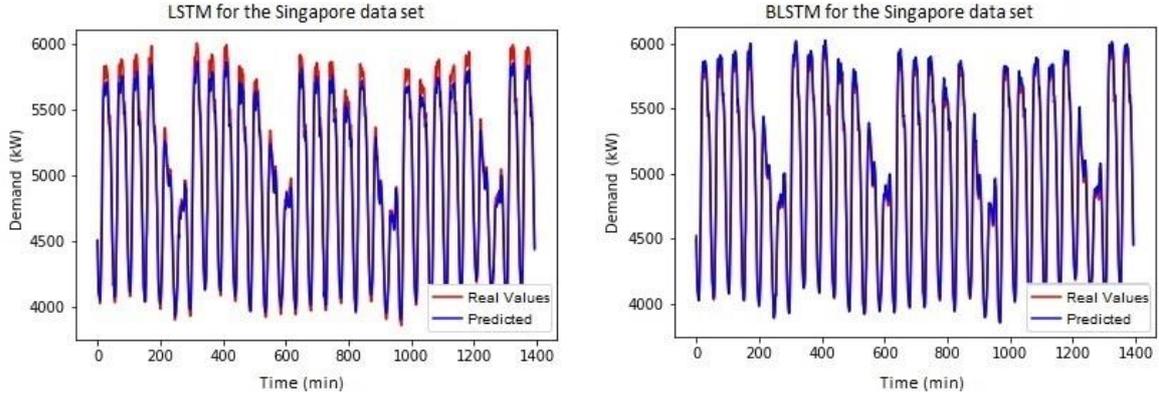

**Fig. 12.** Typical LSTM and BLSTM predictions for the holdout set (Singapore data set).

*5.2.2 Summary of LSTM and BLSTM results*

Table 12 shows the consolidated results (average NRMSE and $R^2$ *score*) for LSTM and BLSTM networks, applied to the data sets UCI-Household, LABIC-Building, Tetouan-Zones, and Singapore. BLSTM models obtained lower NRMSE averages in all data sets. The statistical tests in order to check the statistical significance of the results will be described in the next subsection.

**Table 12.** Table with consolidated results (average RMSE and $R^2$) for all data sets. Best NRMSE averages in bold.

| Network | UCI-Household | | LABIC-Building | | Tetouan-Zones | | Singapore | |
|---|---|---|---|---|---|---|---|---|
| | NRMSE | R² | NRMSE | R² | NRMSE | R² | NRMSE | R² |
| LSTM | 0.070 | 69.5 | 0.032 | 97.8 | 0.016 | 99.6 | 0.017 | 99.7 |
| BLSTM | **0.067** | 72.6 | **0.031** | 97.9 | **0.012** | 99.8 | **0.016** | 99.7 |

**5.3 Statistical Tests**

In order to verify if there is statistical significance of the results, the Friedman test was applied considering the average NRMSE obtained by LSTM and BLSTM regarding all of the four data sets (Table 11).

According to Friedman test there is statistically significance difference between LSTM and BLSTM (p = 0.0455). Nemenyi test also indicated statistically significant difference (p = 0.046). According to such results, BLSTM outperformed LSTM for the time series prediction of electric power consumption.

## 5.4 TS-CV Execution Times

The TS-CV execution times (with $k = 10$) for obtaining the models trained are shown in Table 13. The execution times may be compared with Fig. 13 a comparison of the values. Using the hyperparameters shown in Table 2, BLSTM's TS-CV took more time than LSTM's.

**Table 13.** LSTM and BLSTM's execution times (in hours) for the TS-CV of each data set (k = 10).

| Networks | UCI-Household | LABIC-Building | Tetouan-Zones | Singapore |
|---|---|---|---|---|
|  | Time (h) | Time (h) | Time (h) | Time (h) |
| LSTM | 8.75 | 1.58 | 2.28 | 0.97 |
| BLSTM | 17.76 | 2.58 | 3.93 | 1.63 |

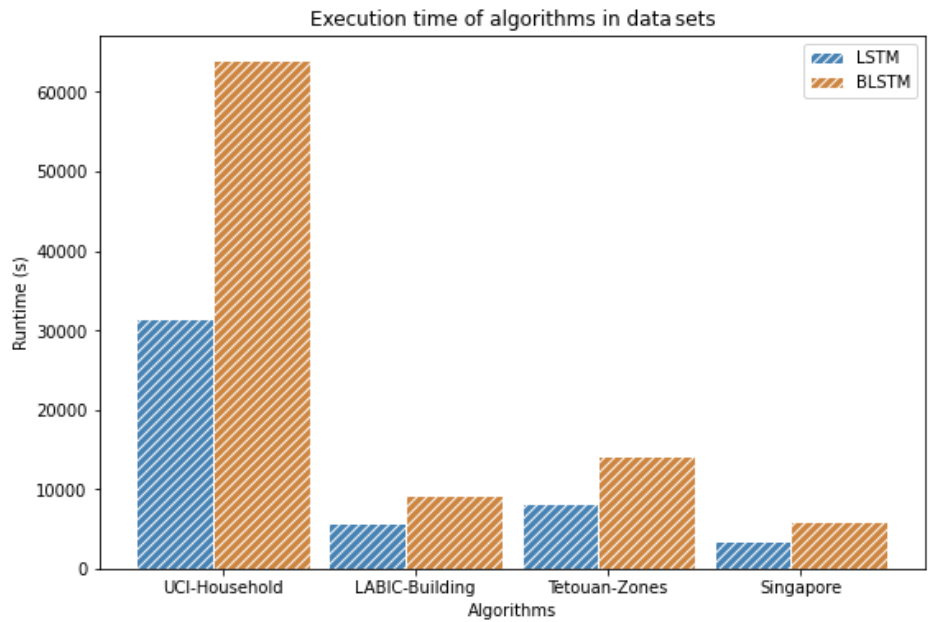

**Fig. 13.** LSTM and BLSTM's execution times for all data sets.

## 6. Discussion

In the present work the LSTM and BLSTM models generated by TS-CV were evaluated in four electric energy consumption holdout data sets. Each data set represents a different scales and consumption characteristics (household, building, city zones, and country). Also, a statistical comparison of the results was performed with the Friedman and Nemenyi tests, indicating that BLSTM outperformed LSTM statistically.

Such BLSTM results corroborate the results obtained for example by Siami-Namini et al. (2019), who compared LSTM and BLSTM performances for TS prediction, and BLSTM obtained lower RMSE values.

Another comparison is reported by Fang and Yuan (2019) which evaluated DNN models for TS forecasting using LSTM and BLSTM. In the same way, their results pointed to better BLSTM performance regarding the metrics MAE, MAPE, and RMSE.

Rhif et al. (2020) also compared LSTM and BLSTM for TS prediction. The authors report the metrics RMSE and $R^2$, and BLSTM was the best DNN in the tests. In the present work, we also contribute reporting statistical evidence for data sets in different scales, supporting that BLSTM outperforms LSTM models.

Regarding the execution times, the higher amount of time necessary for training BLSM (approximately 2.0, 1.6, 1.7, and 1.7 times higher than for LSTM, respectively for each data set) compensates for the results of having higher quality prediction results statistically confirmed.

## 7. Conclusion

Nowadays energy efficiency is a central problem both in economic and environmental perspectives. New technologies are to be tested for monitoring and predicting electric energy, aiming to reduce the consumption through concrete actions, as highlighted by Gardner and Stern (2002).

Thus, a comparison between DNNs for electric consumption time series prediction becomes important in such scenario. LSTM and BLSTM are prominent DNN architectures that are compared in the present work. The statistical comparison of models generated with different lengths of points (generated by TS-CV) and applied to holdout sets is a way of demonstrating their robustness, also taking into account that the TSs were selected considering different scales and characteristics (household, building, city zones and country scales).

As a result BLSTM outperformed LSTM models for energy consumption time series prediction with statistical significance, despite the higher time necessary for training, which compensates for the better results.

**Acknowledgements**

D.G.S. and A.A.M.M. acknowledge Brazilian National Council for Scientific and Technological Development (CNPq) for financial support.

**Data availability**

The dataset LABIC-Building Data Set used in this manuscript can be found at:
https://www.kaggle.com/datasets/daviguimaraes/labic-building-data-set

Conference on Electrical Engineering and Computer Science (EECS), Athens, Greece, 28–30, December; pp. 22–25. https://doi.org/10.1109/EECS49779.2019.00018.

Schmidhuber, J., 1992. **A fixed size storage O (n³) time complexity learning algorithm for fully recurrent continually running networks.** Neural Computation, v. 4, n. 2, p. 243-248. https://doi.org/10.1162/neco.1992.4.2.243.

Schmidhuber, J., 2015. **Deep learning in neural networks: An overview.** Neural networks, v. 61, p. 85-117. https://doi.org/10.1016/j.neunet.2014.09.003.

Schuster, M. and Paliwal K. K., 1997. **Bidirectional recurrent neural networks.** in IEEE Transactions on Signal Processing, vol. 45, no. 11, pp. 2673-2681, Nov. doi: 10.1109/78.650093. https://doi.org/10.1109/78.650093.

Serpanos, D. and Wolf, M., 2018. **Internet-of-Things (IoT) Systems.** Cham: Springer International Publishing. https://doi.org/10.1007/978-3-319-69715-4_5.

Shaqour, A. et al., 2022. **Electrical demand aggregation effects on the performance of deep learning-based short-term load forecasting of a residential building.** Energy and AI, v. 8, p. 100141, 2022. https://doi.org/10.1016/j.egyai.2022.100141.

Sharfuddin, A. A., Tihami, M. N., & Islam, M. S., 2018. **A deep recurrent neural network with BLSTM model for sentiment classification**. International conference on bangla speech and language processing (ICBSLP), IEEE, pp. 1-4. https://doi.org/10.1109/ICBSLP.2018.8554396.

Shin, S.-Y. and Woo, H.-G., 2022. **Energy Consumption Forecasting in Korea Using Machine Learning Algorithms.** Energies, v. 15, n. 13, p. 4880. https://doi.org/10.3390/en15134880.

Siami-Namini, S., Tavakoli, N., & Namin, A. S., 2019. **The performance of LSTM and BLSTM in forecasting time series**. In: 2019 IEEE International Conference on Big Data (Big Data). IEEE, 2019. p. 3285-3292. https://doi.org/10.1109/BigData47090.2019.9005997.

Singh, A. P. et al. **Tetuan City power consumption**. Distribution Network Station of Tetouan city in Morocco. 2018. Retrieved [Date Retrieved] from https://www.kaggle.com/datasets/gmkeshav/tetuan-city-power-consumption (accessed 22 Jun 2022).

Zhao, Y. et al., 2018. **Applying deep bidirectional LSTM and mixture density network for basketball trajectory prediction**. Optik, v. 158, p. 266-272. https://doi.org/10.1016/j.ijleo.2017.12.038.38